\documentclass[document]{llncs}
\usepackage{graphicx}
\usepackage{comment}
\usepackage{amsmath,amssymb} % define this before the line numbering.
\usepackage{color}

\usepackage[width=122mm,left=12mm,paperwidth=146mm,height=193mm,top=12mm,paperheight=217mm]{geometry}

\usepackage{multirow}
\usepackage[table,xcdraw]{xcolor}
\newcommand{\pred}{ZAP}

\begin{document}
% \linenumbers
\pagestyle{headings}
\mainmatter
\def\ECCVSubNumber{1046}  % Insert your submission number here

\title{Thanks for Nothing:\\Predicting Zero-Valued Activations with Lightweight Convolutional Neural Networks} % Replace with your title

\begin{comment}
\titlerunning{ECCV-20 submission ID \ECCVSubNumber} 
\authorrunning{ECCV-20 submission ID \ECCVSubNumber} 
\author{Anonymous ECCV submission}
\institute{Paper ID \ECCVSubNumber}
\end{comment}
%******************

%\begin{comment}
\titlerunning{Thanks for Nothing: Predicting Zero-Valued Activations}
% \orcidID{0000-1111-2222-3333}
\author{Gil Shomron\inst{1} \and
Ron Banner\inst{2} \and
Moran Shkolnik\inst{1,2} \and
Uri Weiser\inst{1}}
\authorrunning{G. Shomron et al.}
\institute{Faculty of Electrical Engineering --- Technion, Haifa, Israel \\
\email{\{gilsho@campus, uri.weiser@ee\}.technion.ac.il} \and
Habana Labs --- An Intel Company, Caesarea, Israel \\
\email{\{rbanner, mshkolnik\}@habana.ai}
}
%\end{comment}
%******************
\maketitle

\begin{abstract}
Convolutional neural networks (CNNs) introduce state-of-the-art results for various tasks with the price of high computational demands.
Inspired by the observation that spatial correlation exists in CNN output feature maps (ofms), we propose a method to dynamically predict whether ofm activations are zero-valued or not according to their neighboring activation values, thereby avoiding zero-valued activations and reducing the number of convolution operations.
We implement the zero activation predictor (\pred{}) with a lightweight CNN, which imposes negligible overheads and is easy to deploy on existing models.
\pred{}s are trained by mimicking hidden layer ouputs; thereby, enabling a parallel and label-free training.
Furthermore, without retraining, each \pred{} can be tuned to a different operating point trading accuracy for MAC reduction.
%For example, using VGG-16 and the ILSVRC-2012 dataset, two different operating points achieve a reduction of 20\% and 30\% multiply-accumulate (MAC) operations with top-1/top-5 accuracy degradation of 0.1\%/0.04\% and 1.3\%/0.7\% without fine-tuning of the entire model, respectively.
%Considering one-epoch fine-tuning, 45\% MAC operations may be reduced with 1.3\%/0.7\% accuracy degradation.

\end{abstract}

\section{Introduction}

In the past decade, convolutional neural networks (CNNs) have been adopted for numerous applications \cite{silver2016mastering}\cite{sainath2013deep}\cite{levine2016end}, introducing state-of-the-art results.
Despite being widely used, CNNs involve a considerable amount of computations.
For example, classification of a 224x224 colored image requires billions of multiply-accumulate (MAC) operations \cite{canziani2016analysis}\cite{sze2017efficient}.
Such computational loads have many implications, from execution time to power and energy consumption of the underlying hardware.

\begin{figure}[t] 
    \centering
	\includegraphics[height=4cm]{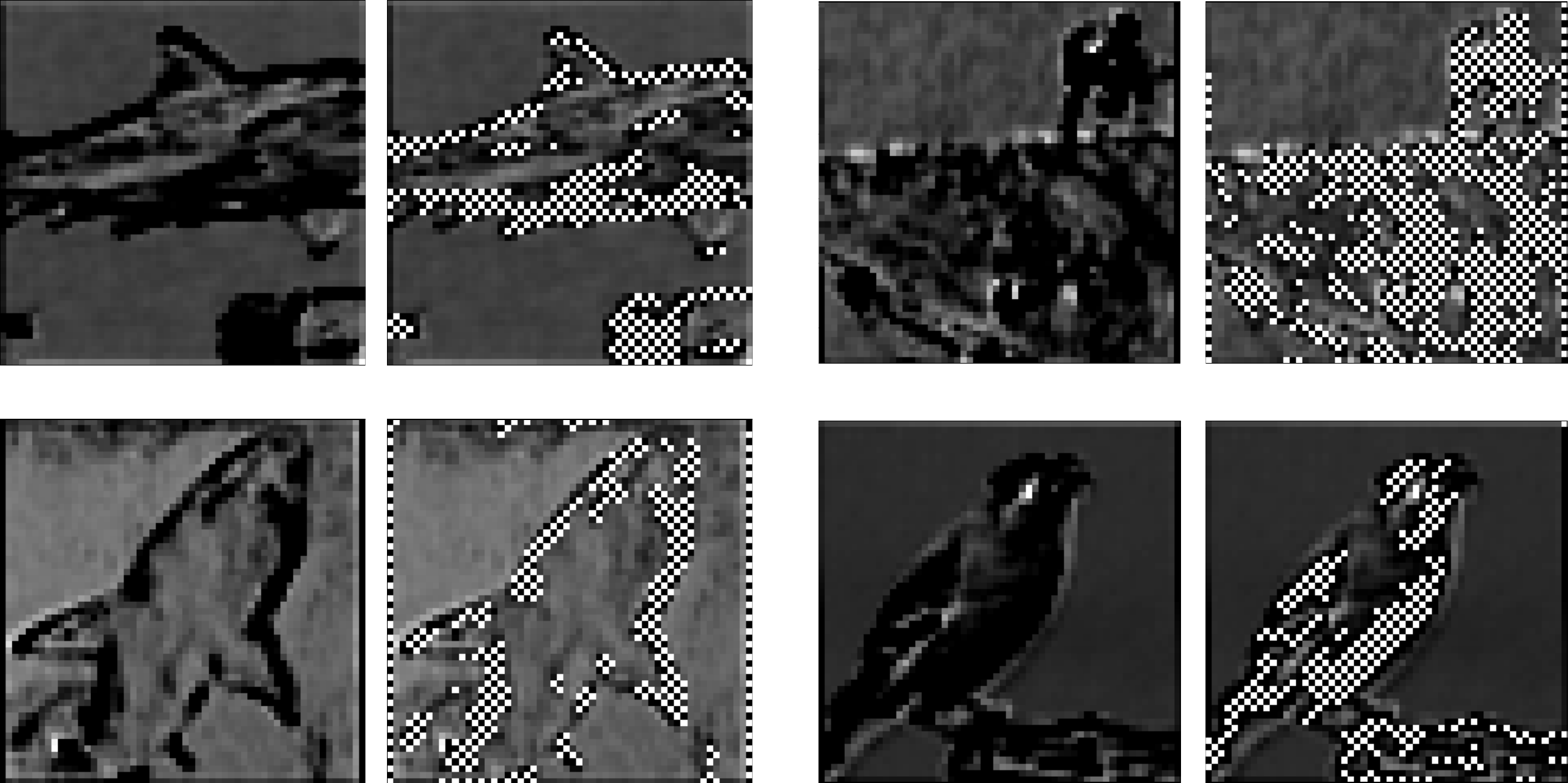}
	
	\caption{Exploiting spatial correlation for zero-value prediction of ofm activations with CNN-based predictor.
			 Bright white pixels represent a predicted zero-valued ofm activation.}
	\label{fig:intro:examples}
\end{figure}

CNN output feature maps (ofms) have been observed to exhibit spatial correlation, i.e., adjacent ofm activations share close values \cite{mahmoud2018diffy}\cite{shomron2019spatial}.
This observation is particularly true for zero-valued activations, as it is common practice to use the ReLU activation function \cite{nair2010rectified}, which squeezes all negative values to zero.
If it were possible to predict which of the convolution operations will result in a negative value, they could be skipped, their corresponding ofm activations could be set to zero, and many multiply-accumulate (MAC) operations could be saved.%, which in turn may save energy.

Prediction mechanisms are at the heart of many general-purpose processors (GPPs), leveraging unique application characteristics, such as code semantics and temporal locality, to predict branch prediction outcomes and future memory accesses, for example.
Prediction mechanisms may similarly be employed for CNNs.
In this paper, we propose a prediction method for CNNs that dynamically classifies ofm activations as zero-valued or non-zero-valued by leveraging the spatial correlation of ofms.
The zero activation predictor (\pred{}) works in three-steps:
first, only a portion of the ofm is fully computed;
then, the remaining activations are classified as zero-valued or non-zero-valued using a lightweight convolutional neural network;
and finally, the predicted non-zero-valued activations are computed while the zero-valued activations are skipped, thereby saving entire convolution operations (Figure~\ref{fig:intro:examples}).
\pred{} imposes negligible overheads in terms of computations and memory footprint, it may be plugged into pretrained models, it is trained quickly, in parallel, and does not require labeled data.
%In addition, \pred{}'s speculation level is tunable, enabling different operating options according to existing constraints.

\pred{} may be considered as a CNN-based, dynamic, unstructured, and magnitude-based ofm activations pruning strategy.
However, as opposed to many pruning techniques, \pred{} is tunable on-the-fly.
Therefore, \pred{} captures a wide range of operating points, trading accuracy for MAC savings.
For example, by strictly focusing on zero-valued ofm activations, \pred{} can capture a range of operating points that does not require retraining.
\pred{} is also capable of pruning when set to high speculation levels, as more mispredictions of non-zero-valued activations as zero-valued activations take place.
Interestingly, we observe that these mispredictions usually occur with small activation values, which is practically a magnitude-based pruning strategy.

%\pred{} may be considered a dynamic, unstructured, and magnitude-based ofm activations pruning strategy.
%Pruning typically includes two stages: network trimming and retraining.
%First, weights \cite{han2015deep}\cite{li2016pruning} and activations \cite{lin2017runtime}\cite{he2017channel} are classified as insignificant according to some criterion and are %discarded, followed by retraining phases to recoup the loss in accuracy.
%Increased model sparsity, when accompanied with the appropriate hardware \cite{albericio2016cnvlutin}\cite{han2016eie}\cite{parashar2017scnn}\cite{shomron2019smt}, leads to a decrease in MAC operations.
%Since our method focuses strictly on the zero-valued ofm activations, \pred{} can capture a range of operation points that does not require retraining.
%That said, \pred{} is also capable of pruning when set to high speculation levels, as more mispredictions of non-zero-valued activations as zero-valued activations take place.
%Interestingly, we observe that these mispredictions usually occur with small activation values, which is practically a magnitude-based pruning strategy.

This paper makes the following contributions: 
\begin{itemize}
	\item \textbf{Zero activation predictor (ZAP).} We introduce a dynamic, easy to deploy, CNN-based zero-value prediction method that exploits the spatial correlation of output feature map activations. Compared with conventional convolution layers, our method imposes negligible overheads in terms of both computations and parameters.
	\item \textbf{Trade-off control.} We estimate the model's entire accuracy-to-savings trade-off curve with mostly local statistics gathered by each predictor. This provides a projection of the model and the predictor performance for any operating point.
	\item \textbf{Accuracy to error linearity.} We show, both analytically and empirically, that the entire model accuracy is linear with the sum of local misprediction errors, assuming they are sufficiently small.
	\item \textbf{Local tuning given global constraints.} We consider layers variability by optimizing each \pred{} to minimize the model MAC operations subject to a global error budget.

\end{itemize}

\section{\pred{}: Zero Activation Predictor}
In this section, we describe our prediction method and its savings potential.
We analyze its overheads in terms of computational cost and memory footprint and show that both are negligible compared with conventional convolution layers.

\subsection{Preliminary}
A convolution layer consists of a triplet $\langle X_{i},X_{o},W \rangle$, where $X_{i}$ and $X_{o}$ correspond to the input and output activation tensors, respectively, and $W$ corresponds to the set of convolution kernels.
Each activation tensor is a three-dimensional tensor of size $w \times h \times c$, where $w$, $h$ and $c$ represent the tensor width, height, and depth, respectively.
For convenience, the dimensions of $X_{i}$ and $X_{o}$ are denoted by $[w_{i} \times h_{i} \times c_{i}]$ and $[w_{o} \times h_{o} \times c_{o}]$.
Finally, the set of convolution kernels $W$ is denoted by a four-dimensional tensor $[k \times k \times c_i \times c_o]$, where $k$ is the filter width and height.
The filter width and height are not necessarily equal, but it is common practice in most conventional CNNs to take them a such.
Given the above notations, each output activation value $X_o[x,y,z]$ is computed from the input tensor $X_i$ and weights $W$ as follows:
\begin{equation}
X_o(x,y,z) = \sum_{i,j=0}^{k-1} \sum_{c=0}^{c_i-1} X_i[x+i, y+j, c] \cdot W[i,j,c,z],
\label{eq:conv}
\end{equation}
where the bias term is omitted for simplicity's sake.
We use the above notations throughout this paper.

\subsection{Three-Step Method}
\label{sec:three_step-method}
%Conventional convolution layers compute their ofm by applying Equation~(\ref{eq:conv}) for each $[x, y, z]$.
%It has been observed that ofms accommodate many zero-valued activations due to the widespread usage of the ReLU activation function.
%For example, in Figure~\ref{fig:resnet-18:sparsity} we present the relative portion of zero-valued activations of ResNet-18 \cite{he2016deep} with the 2012 ImageNet large scale visual recognition challenge (ILSVRC-2012) dataset \cite{russakovsky2015imagenet} --- overall, almost 50\% of the ofm activations are zero-valued; therefore, 50\% of the MAC operations may be potentially skipped. 
%Moreover, ofm activations exhibit spatial correlation \cite{mahmoud2018diffy}\cite{shomron2019spatial}, meaning that a group of activation values testifies about other adjacent activation values.
Conventional convolution layers compute their ofm by applying Equation~(\ref{eq:conv}) for each $[x, y, z]$.
It has been observed that ofms accommodate many zero-valued activations due to the widespread usage of the ReLU activation function \cite{parashar2017scnn}\cite{sze2017efficient}\cite{albericio2016cnvlutin}.
For example, with ResNet-18 \cite{he2016deep}, almost 50\% of the ofm activations are zero-valued; therefore, 50\% of the MAC operations may be potentially skipped. 
Moreover, ofm activations exhibit spatial correlation \cite{mahmoud2018diffy}\cite{shomron2019spatial}, meaning that a group of activation values testifies about other adjacent activation values.

\begin{figure}[t] 
    \centering

	\includegraphics[width=1.0\textwidth]{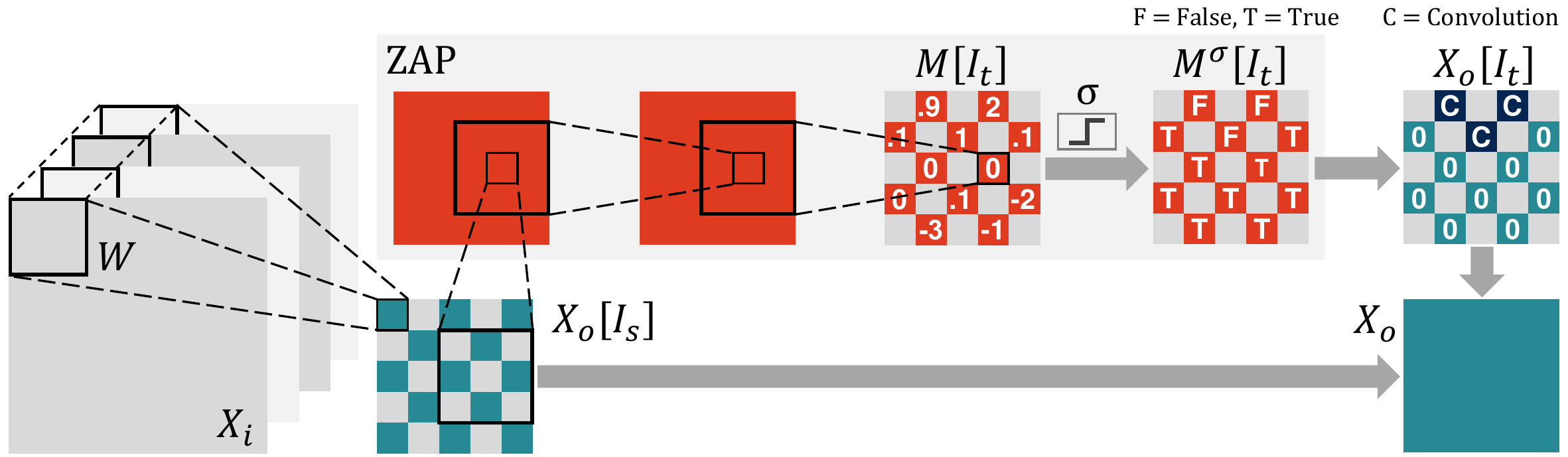}

	\caption{Illustration of a single ofm channel convolution with \pred{}.
	$X_o[I_s]$ is computed based on a pre-defined pattern.
	$X_o[I_s]$ is then subjected to \pred{}, which produces a prediction map $M[I_t]$, followed by a thresholding step to form the binary prediction map $M^{\sigma}[I_t]$.
	$X_o[I_t]$ is created according to $M^{\sigma}[I_t]$ --- a portion of the $I_t$ ofm activations are predicted as zero-valued and skipped, whereas the rest are computed using Equation~(\ref{eq:conv}).
	%$I_t$ ofm activations are either predicted as zero-valued and skipped, or computed using Equation~(\ref{eq:conv}).
	%$\hat{X}_o$ is a combination of $\tilde{X}_o$ and complementary convolution operations that are taking place at $(x,y,z)$ coordinates where $\tilde{X}^c_o(x,y,z) > \sigma$.
	}
	\label{fig:pred_mech}
\end{figure}

We suggest a three-step dynamic prediction mechanism that locates zero-valued ofm activations by exploiting the spatial correlation inherent in CNNs to potentially skip them and reduce the computational burden of the model.
Given an ofm, $X_o$, we divide its indices into two subsets $(I_s,I_t)$ according to a pre-defined pattern.
Then, the following three steps are carried out (as illustrated in Figure~\ref{fig:pred_mech}):
(i) the values that belong to indices $I_s$ are fully computed in the conventional manner, resulting in a sparse ofm, $X_o[I_s]$;
(ii) this partial ofm ($X_o[I_s]$) is passed through our predictor to yield a binary map, $M^{\sigma}[I_t]$, which is used to predict the zero values in $X_o[I_t]$; and
(iii) all values predicted to be non-zero by $M^{\sigma}[I_t]$ are computed.
We describe this process in detail next.

\begin{figure}[t] 
    \centering
	\includegraphics[height=2.2cm]{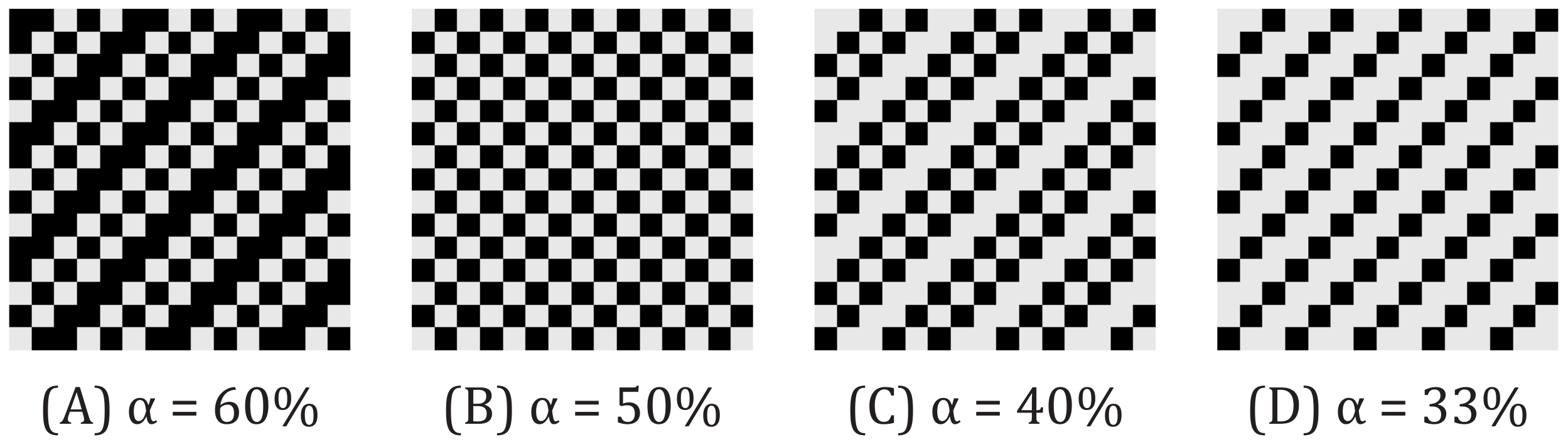}
	
	\caption{Partial ofm convolution patterns.
		     Black pixels represent computed activations.}
	\label{fig:pre-comp_pats}
\end{figure}

\textbf{Computation pattern.}
$X_o[I_s]$ is computed based on a pre-defined pattern as depicted in Figure~\ref{fig:pre-comp_pats}.
Since our predictor exploits spatial correlation, we partially compute the ofm so that activations that are not computed in the first step will reside next to a number of computed activations.
We use $\alpha$ to denote the ratio between the number of activations computed in the partial ofm and the ofm dimensions.
This may be formally formulated as $\alpha \equiv \frac{|I_s|}{w_o h_o c_o}$, where $|I_s|$ is the set cardinality.

By decreasing $\alpha$, less ofm activations are computed prior to the prediction, which may potentially lead to the saving of more operations.
For example, for $\alpha=40\%$, 60\% of the activations may potentially be saved.
Less prior data about the ofm may, however, lead to higher misprediction rates, which in turn may decrease model accuracy.

\textbf{Prediction process via lightweight CNN.}
Given the partial ofm, $X_o[I_s]$, our goal is to produce an output computation mask that predicts which of the remaining activations are zero-valued and may be skipped.
Recall that ofm activations are originally computed using Equation~(\ref{eq:conv}) and so the prediction process must involve less MAC operations than the original convolution operation with as minimal memory footprint as possible.

We use a CNN to implement \pred{}.
We exploit the spatial correlation inherent in CNNs \cite{mahmoud2018diffy}\cite{shomron2019spatial}\cite{kim2018mosaic}\cite{figurnov2016perforatedcnns} and use only depthwise convolution (DW-CONV) layers \cite{howard2017mobilenets}, i.e., only spatial filters with no depth ($k \times k \times 1 \times c_o$).
As such, we obtain a lightweight model in terms of both MAC operations and parameters (further analyzed in Section~\ref{sec:overhead_analysis}).
Our CNN comprises a 3x3 DW-CONV layer followed by a batch normalization (BN) layer \cite{ioffe2015batch} followed by ReLU, twice.
DW-CONV layer padding and stride are both defined as 1 to achieve equal dimensions throughout the CNN predictor.
During training, the last ReLU is capped at 1 \cite{krizhevsky2010convolutional}, whereas during inference we discard the last ReLU activation and use a threshold.

\textbf{Thresholding.}
Altough \pred{} is trained to output a binary classifier (described in Section~\ref{sec:training}), \pred{} naturally outputs $M[I_t]$, which is not a strict binary mask but rather a range of values that corresponds to a prediction confidence \cite{geifman2017selective}\cite{yazdani2018dark} of whether the ofm activation is zero-valued or not.
To binarize \pred{}'s output, we define a threshold $\sigma$ that sets the level of prediction confidence; therefore, $M^{\sigma}[I_t] = M[I_t] > \sigma$ (boolean operation).

According to $M^{\sigma}[I_t]$, part of the ofm activations in $X_o[I_t]$ are predicted to be non-zero-valued and are computed, whereas the others are predicted to be zero-valued and are skipped.
When an ofm activation is predicted to be zero-valued, two types of misprediction may occur.
First, an actual zero-valued activation may be predicted as a non-zero-valued activation and so redundant MAC operations may take place.
Second, an actual non-zero-valued activation may be predicted as zero.
The latter misprediction increases the model error, potentially decreasing model accuracy.

The motivation behind \pred{} is clear --- reducing computations by skipping convolution operations.
Its impact on the model accuracy, number of computations, and amount of parameters has yet, however, to be discussed.
We next discuss the overhead of \pred{}, and in Section~\ref{sec:experiments}, we show empirically how \pred{} affects the accuracy of different model architectures.

\subsection{Overhead Analysis}
\label{sec:overhead_analysis}
\pred{} is a CNN by itself, which means that it introduces additional computations and parameters.
To be beneficial, it must execute less operations per ofm activation than the original convolution operation.
A conventional convolution layer requires $k^2 c_i$ MAC operations per ofm activation.
On the other hand, the number of MAC operations required by a two-layered DW-CONV \pred{} for a single ofm activation is $K^2$, where $K$ is the filter width and height.
Note that \pred{}'s first layer needs only to consider $|I_s|$ values for computation since, according to the pre-defined pattern, the remaining $|I_t|$ values are zero; the second layer needs only to compute $|I_t|$ ofm activations.

Compared with a standard convolution operation, and for the case of $K=k$,
\begin{equation}
\frac{\text{\pred{} ops.}}{\text{standard convolution ops.}} = \frac{1}{c_i} \,.
\label{eq:overhead}
\end{equation}
$c_i$ is usually greater than $10^2$ \cite{he2016deep}\cite{simonyan2015very}\cite{krizhevsky2012imagenet}, in which case $1 / c_i \approx 0$ and \pred{} overhead is, therefore, negligible.

Regarding the parameters, a conventional convolution layer requires $k^2 c_i c_o$ parameters and a two-layered DW-CONV requires $2 K^2 c_o$ parameters.
Given $c_i$ conventional sizes and $K=k$, \pred{}'s memory footprint is negligible as well.

\section{Trade-Off Control}
The threshold hyperparameter, $\sigma$, represents the prediction confidence of whether an ofm activation is zero or non-zero.
Users should, however, address the accuracy and MAC reduction terms rather than using $\sigma$, since it is not clear how a specific $\sigma$ value affects accuracy and savings.
In this section, we show that, given some statistics, we can estimate the entire model accuracy and predictor MAC savings, thereby avoiding the need to address the threshold value directly and providing the user with an accuracy-MAC savings trade-off control ``knob''.

\subsection{Accuracy, MAC Savings, and Threshold}
%\label{sec:acc-err-linear}
%Not only the MAC operations may be estimated for different thresholds, but also the model accuracy, again when assuming low noise.
\textbf{Accuracy and threshold.} Consider a DNN with $L$ layers.
Each layer $i$ comprises weights $w_i$, an ifm $x_i$, and an ofm $y_i$.
When predictions are avoided, layer $i+1$ input, $x_{i+1}$, is given by
\begin{equation}
    x_{i+1} = y_i = \max(x_i w_i,0) \,,
\end{equation}
where $\max(\cdot, 0)$ is the ReLU activation function.

Our predictor is not perfect and may zero out non-zero elements with a small probability $\varepsilon_i$.
Therefore, non-zero inputs to layer $i+1$ have a probability $\varepsilon_i$ of becoming zero and a probability $1-\varepsilon$ of remaining $y_i$.
Using $y^{\pi}_{i}$ to denote the prediction of output $y_i$, we obtain the following error in the expected activation:
\begin{equation}
    \label{eq:expected_error}
    E(y^{\pi}_{i}) =  (1-\varepsilon_i)\cdot E(y_{i}) \,.
\end{equation}
In other words, the prediction at each layer $i$ introduces a multiplicative error of $(1-\varepsilon_i)$ with respect to the true activation.
This multiplicative error builds up across layers. For an $L$-layer network, the expected network outputs are scaled down with respect to the true network outputs as follows: 
\begin{equation}
    \text{Scale Error}= \prod_{i=1}^{L} (1-\varepsilon_i) \,.
\end{equation}
Note that for a sufficiently small $\varepsilon$, $1-\varepsilon = e^{-\varepsilon}$.
Therefore, assuming sufficiently small misprediction probabilities, $\{\varepsilon_i\}$, we obtain the following expression 
\begin{equation}
\begin{aligned}
    \text{Scale Error} &= \prod_{i=1}^{L} (1-\varepsilon_i)\approx \prod_{i=1}^{L} e^{-\varepsilon_i} \\
    &=e^{-\sum_{i=1}^{L}\varepsilon_i}
     \approx  1-\sum_{i=1}^{L}\varepsilon_i \,,
\end{aligned}
\label{eq:bias}
\end{equation}
where the approximations can easily be extracted from a Taylor expansion to $e^{-x}$.
Equation~(\ref{eq:bias}) shows that the final error due to small mispredictions accumulates along the network in a \textit{linear} fashion when the errors due to mispredictions are small enough.

\begin{figure*}[t]
	\centering
	\includegraphics[width=0.95\textwidth]{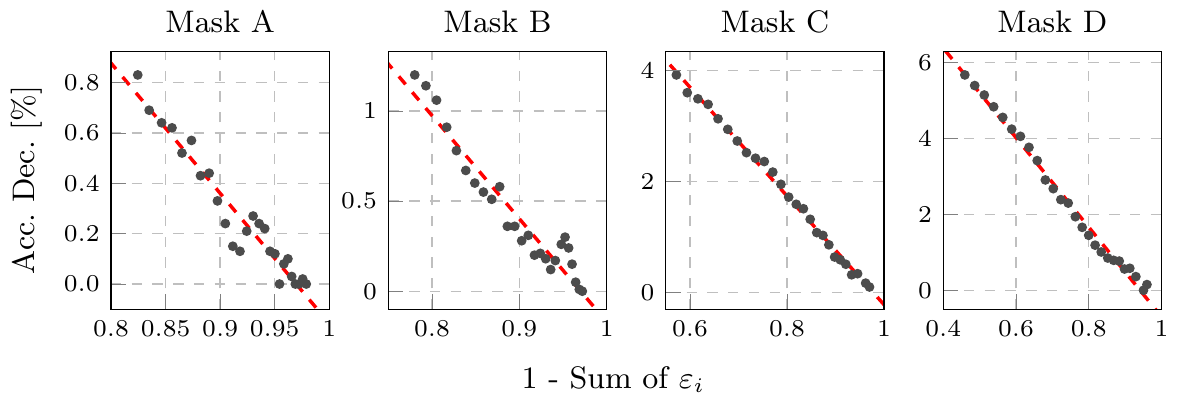}	
	\caption{AlexNet + CIFAR-100 top-1 accuracy decrease as a function of $1 - \sum_{i=1}^L \varepsilon_i(\sigma)$.
			 Each dot represents a measurement with a different threshold: the leftmost and rightmost thresholds are 0 and 0.5, respectively.
			 Measurements were taken in steps of 0.02.}
	\label{fig:cifar100:err-acc}
\end{figure*}

Denoting the output of layer $i$ for a threshold $\sigma$ by $y_i^{\pi,\sigma}$, the error is associated with a threshold $\sigma$ as follows:
\begin{equation}
\begin{aligned}
\text{Scale Error}\,(\sigma) & \approx 1 - \sum_{i=1}^L \varepsilon_i(\sigma) \\
 = 1 - & \sum_{i=1}^L \left( 1 - \frac{\sum_{x,y,z} y_i^{\pi,\sigma}[x,y,z]}{\sum_{x,y,z} y_i[x,y,z]} \right) \,,
\end{aligned}
\label{eq:scale_error}
\end{equation}
where Equation~(\ref{eq:expected_error}) is used for the last transition.
Figure~\ref{fig:cifar100:err-acc} shows that this analytical observation is in good agreement with our empirical results.  

Ideally, we would like to have a scale error that is as close as possible to 1 so as to avoid shifting the network statistics from the learned distribution.
When the scale error is 1, network outputs remain unchanged and no accuracy degradation associated with mispredictions occurs.
Yet, as $\sigma$ increases, more ofm activations are predicted as zero-valued, decreasing the scale error below 1, as exhibited by Equation~(\ref{eq:scale_error}).
This scale error decreases monotonically with $\sigma$ and introduces an inverse mapping, which enables us to define a threshold value for any desired accuracy degradation.   

\textbf{MAC savings and threshold.} Recall that $M[I_t]$ represents the collection of values predicted by \pred{} for a given layer before applying the threshold.
%For the sake of readability
For readability, assume $m(x)$ is the probability density function of $M[I_t]$ values (i.e., continuous).
Then, the number of ofm activations predicted as zeros relative to $I_t$ can be expressed as follows:
\begin{equation}
\text{Zero Prediction Rate}\,(\sigma) = \int_{-\infty}^{\sigma} m(x) dx \,.
\label{eq:pred_rate_integral}
\end{equation}
To achieve the actual MAC reduction in layer $i$, the layer dimensions, $\alpha$, and \pred{} overhead should be considered.
Clearly, the total MAC reduction equals the sum of the contributions from all layers.
Note that the zero prediction rate, and therefore the MAC reduction, increases monotonically, and for a certain range it may be considered a strictly monotonic function.
Inverse mapping from MAC reduction to $\sigma$ is, therefore, possible.%, as depicted in Figure~\ref{fig:inverse_function}.

%\begin{figure}[t]
%	\centering
%	\includegraphics[width=8.4cm]{./imgs/mac_sigma.pdf}
%	\caption{The transformation process from a prediction values histogram (or distribution) to the threshold versus MAC reduction curve.}
%	\label{fig:inverse_function}
%\end{figure}

\begin{figure}[t] 
    \centering
	\includegraphics[width=0.85\textwidth]{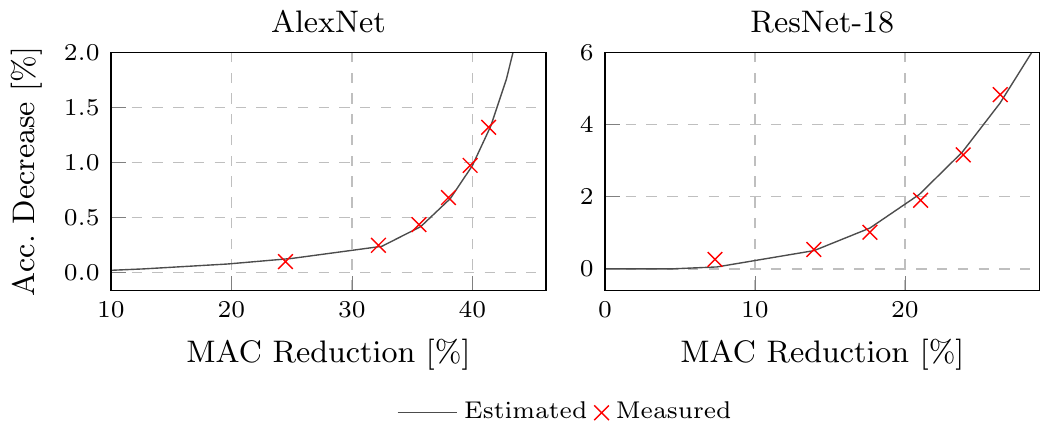}
	\caption{Estimated top-5 accuracy-MAC savings trade-off curve using mask B and ILSVRC-2012 dataset.
		     The measured operating points correspond to thresholds of 0 to 0.5, in steps of 0.1.}
	\label{fig:flow}
\end{figure}

\textbf{Putting it all together.} Given the derivations so far, it is possible to make an \textit{a priori} choice of any operating point, given estimates about the desired accuracy and MAC reduction, without directly dealing with the threshold.
Assume a pre-processing step that consists of collecting statistical information about the prediction values $M[I_t]$ and at least two accuracy measurements of the entire model, to obtain the linear accuracy-error relationship.
With this statistical information, we can effectively estimate the desired operating point, as demonstrated in Figure~\ref{fig:flow}.

\subsection{Non-Uniform Threshold}
\begin{figure}[t] 
    \centering
	\includegraphics[width=0.95\textwidth]{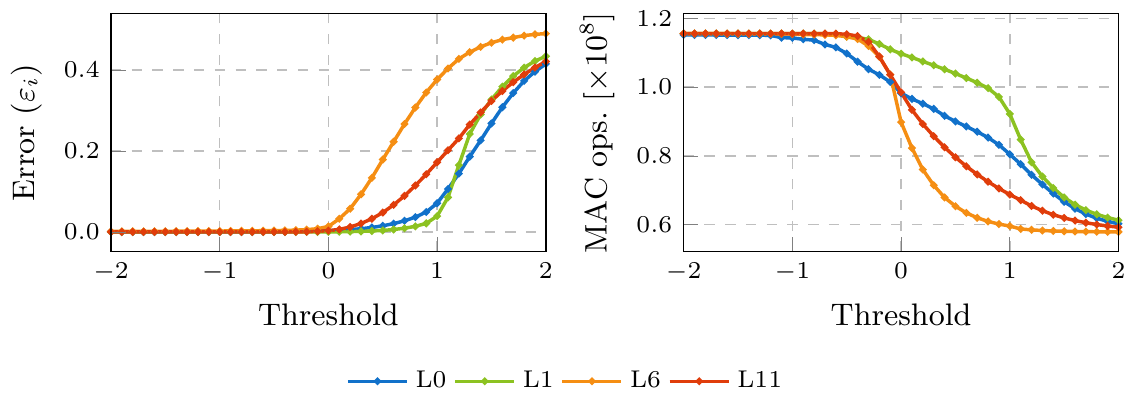}
	\caption{ResNet-18 + ILSVRC-2012 example of different layers behavior in terms of error and MAC operations as a function of \pred{} threshold.}
	\label{fig:err-mac-th}
\end{figure}

Thus far, for the sake of simplicity, $\sigma$ was set uniformly across all layers.
Layers, however, behave differently, and so \pred{} error and savings may differ between layers for a given threshold.
This is evident in Figure~\ref{fig:err-mac-th} in which we present the error and total MAC operations of four layers in ResNet-18.
Notice how the error of layer 6 (L6) increases earlier than the other layers, for example.
Ideally, given $L$ layers, we would like to choose a threshold per layer, $\sigma_i$, to save as many MAC operations as possible, given an error (or accuracy) constraint, $\epsilon$, i.e.,
\begin{equation}
\begin{aligned}
& {\text{minimize}}
& & \sum_{i=1}^L \text{MAC ops.}_i\,(\sigma_i) \\
& \text{subject to}
& & \sum_{i=1}^L \varepsilon_i (\sigma_i) \leq \epsilon \propto \text{Accuracy}.
\end{aligned}
\label{eq:non-linear}
\end{equation}
We use curve fitting to define a parameterized sigmoid function for the error and for the MAC operations of each layer and apply standard non-linear optimization methods to solve Equation~(\ref{eq:non-linear}) (see supplementary material for curve fitting results).
It is worth mentioning that Equation~(\ref{eq:non-linear}) can also be written the other way around, that is, minimizing the error given a computation budget.

\section{Experiments}
\label{sec:experiments}
In this section, we evaluate \pred{} performance in terms of MAC savings and accuracy degradation using various model architectures and datasets, and compare our results to previous work.
%We show how it can be leveraged to estimate the model accuracy and MAC savings for any operating point.

\subsection{\pred{} Training}
\label{sec:training}
\begin{figure}[t] 
    \centering
	\includegraphics[width=0.99\textwidth]{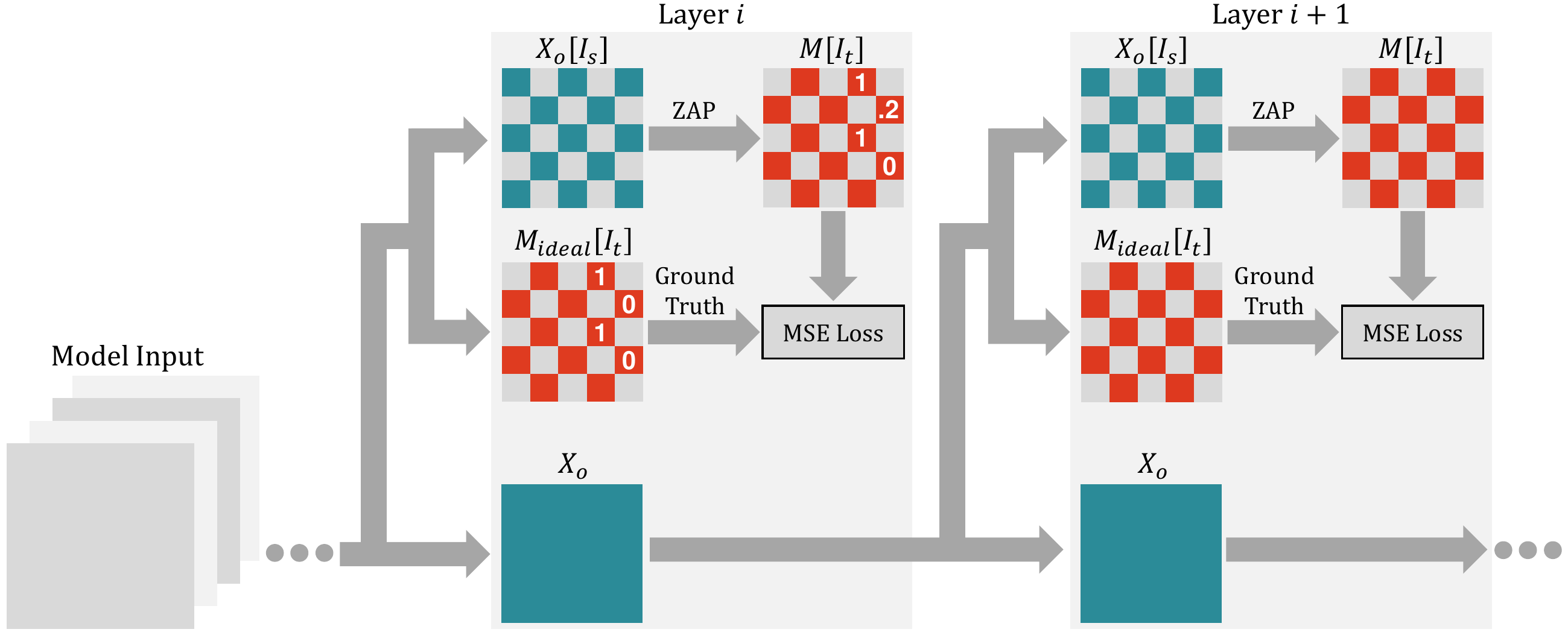}
	\caption{Illustration of \pred{} training. Each \pred{} is trained independently in a teacher-student manner, enabling a parallel label-free training.
		     The ground truth is the original layer ofm after a binary threshold operation (\textgreater 0).
		     The CNN predictor is used with the ReLU activation function capped at 1.}
	\label{fig:mse-train}
\end{figure}

\pred{}s are deployed at each desired convolution layer and are trained independently. %
By training in isolation \cite{he2017channel}\cite{elthakeb2019divide}, \pred{}s can be plugged into a model without altering its architecture and trained parameters and may be trained in parallel.
First, a batch is fed forward bypassing all predictors.
During the feedforward phase, each \pred{} saves a copy of its local input feature map (ifm) and corresponding local ofm.
Then, each \pred{} computes its $M[I_t]$, using its ifm followed by a ReLU activation function which is capped at 1.
The ground truth, $M_{ideal}$, of each \pred{} is its local ofm passed through a zero-threshold boolean operation at indices $I_t$.
Finally, the MSE loss is used to minimize each of the predictor errors, as follows:
\begin{equation}
\min \sum_{(x,y,z) \in I_t} \left( M - M_{ideal}   \right)^2.
\end{equation}
Notice that no labeled data is needed.
\pred{} training is illustrated in Figure~\ref{fig:mse-train}.

\subsection{Experimental Setup}
We evaluated our method using CIFAR-100 \cite{cifardataset} and ILSVRC-2012 \cite{russakovsky2015imagenet} datasets, and AlexNet \cite{krizhevsky2012imagenet}, VGG-16 \cite{simonyan2015very}, and ResNet-18 \cite{he2016deep} CNN architectures.
%PyTorch 1.1.0 \cite{paszke2017automatic} was used as our deep learning framework.
The source code is publicly available\footnote{\url{https://github.com/gilshm/zap}}.

\pred{}s were trained using the Adam~\cite{kingma2014adam} optimizer for 5 epochs.
When the ILSVRC-2012 dataset was used for \pred{}s training, only 32K training set images were used.
After \pred{}s were trained, we recalibrated the running mean and variance of the model BN layers; this is not considered fine-tuning since it does not involve backpropagation.
BN recalibration was noticeably important with ResNet-18, which has multiple BN layers.
When fine-tuning was considered, it was limited to 5 epochs with CIFAR-100 and to 1 epoch with ILSVRC-2012.
%The models were fine-tuned using their original hyperparameters with the smallest learning rate values.
We did not deploy \pred{}s on the first layers of any of the models, since it is not beneficial in terms of potential operation savings.

AlexNet with CIFAR-100 was trained from scratch using the original hyperparameters, achieving top-1 accuracy of 64.4\%.
AlexNet, VGG-16, and ResNet-18 with ILSVRC-2012 were used with the PyTorch pretrained parameters, achieving top-1/top-5 accuracies of 56.5\%/79.1\%, 71.6\%/90.6\%, and 69.8\%/89.1\%, respectively.
%For example, the first AlexNet convolution layer filter dimensions are $3 \times 3 \times 3 \times 64$, meaning the number of MAC operations per ofm activations is 27, which does not leave almost any room for improvement.

We experimented with four different prediction patterns (Figure~\ref{fig:pre-comp_pats}).
The same prediction pattern was used across all layers and channels.
Mixing patterns in layer and channel granularity is an option we leave for future work.

Throughout this section, MAC reduction is defined as $(1 - \text{after}/\text{before})$ and accuracy degradation is defined as $(\text{before} - \text{after})$.
When discussing MAC reduction, we consider only the relative savings from convolution layers.

\begin{figure}[t]
	\centering
	\includegraphics[width=0.95\textwidth]{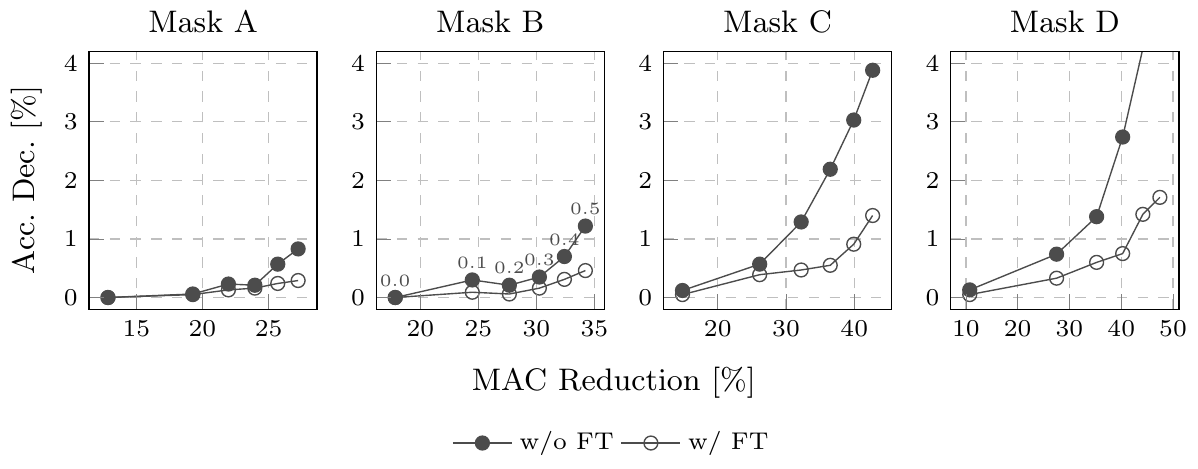}
	\caption{Demonstrating the different operating points of AlexNet + CIFAR-100 with (w/) and without (w/o) fine-tuning (FT).
			 Accuracy measurements are top-1.
 			 Threshold range is set between 0 to 0.5, in steps of 0.1.
			 Each measurement corresponds to a threshold, as presented in ``Mask B'' plot.}
	\label{fig:alexnet-cifar100:accuracy-mac}
\end{figure}

\subsection{CIFAR-100}
\label{sec:exp:cifar}

\textbf{Operating points.}
We demonstrate different operating points as measured with different masks and uniform thresholds across all layers with AlexNet using the CIFAR-100 dataset.
For each operating point, we report the entire model top-1 accuracy degradation and MAC operation savings, with and without fine-tuning (Figure~\ref{fig:alexnet-cifar100:accuracy-mac}).
For example, considering a 1\% top-1 accuracy degradation cap, \pred{} achieves a 32.4\% MAC reduction with 0.7\% accuracy degradation at mask B with a 0.4 threshold.
In addition, by fine-tuning the model, our predictor achieves 40.3\% MAC reduction with 0.8\% accuracy degradation at mask D with a 0.3 threshold.

Masks with greater $\alpha$ values lead to ofms with relatively more computed activations, i.e., larger $|I_s|$.
We observe that these masks show better accuracy results in the low degradation region (e.g., mask A versus mask D at 20\% MAC reduction), whereas for higher MAC reduction requirements, masks with lower $\alpha$ values are preferred.
In order to achieve high MAC reductions with the former masks (for example, mask A), $\sigma$ would have to be cranked up.
Since the prediction potential of these masks is low to begin with (for example, the best-case scenario with mask A is 40\% activations savings with mask A), high thresholds will lead to pruning of relatively significant values and, as a result, to significant accuracy degradation.
On the other hand, for conservative MAC reductions, these masks are preferred, since their speculation levels are lower and prediction confidence is higher.

\begin{figure}[t] 
	\centering
	\includegraphics[width=0.95\textwidth]{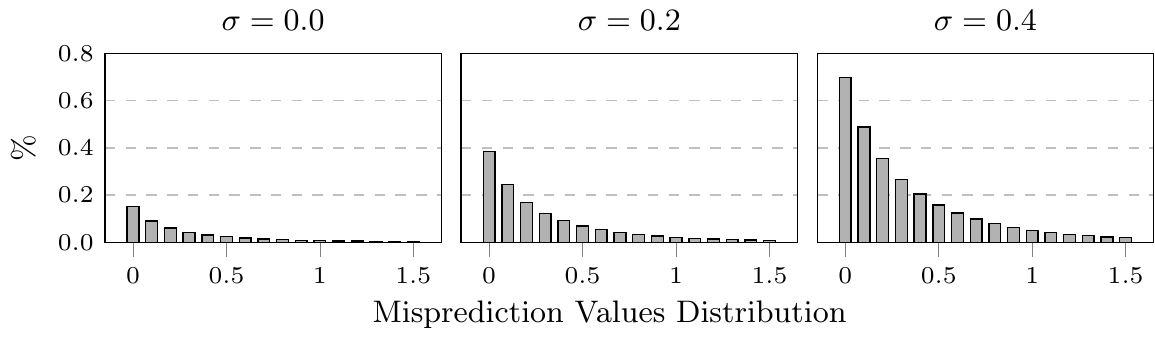}
	\caption{AlexNet + CIFAR-100 histograms of non-zero values that were predicted as zero using mask B and different uniform thresholds.}
	\label{fig:alexnet-cifar100:mispred-hist}
\end{figure}

\textbf{Misprediction breakdown.}
The model accuracy is not solely dependent on \emph{how many} non-zeros were predicted as zero, but also \emph{which} values were zeroed out.
The notion that small values within DNN are relatively ineffectual is at the core of many pruning techniques.
In Figure~\ref{fig:alexnet-cifar100:mispred-hist} we present the mispredicted activation values distribution, normalized to the number of ofm activations for value steps of 0.1.
For example, when using $\sigma=0$, 0.15\% of ofm activations with original values between 0 to 0.1 were zeroed out.
%Another example, for threshold of 0.4 there are 1.0\% of ofm activations with original value between 0.4 to 0.5 that were zeroed out.
Increasing the threshold level increases the number of predicted and mispredicted ofm activations.
It is apparent that insignificant values are more prone to be mispredicted --- it is easier to be mistaken about zeroing out a small value than a large value --- attesting to the correlation between $\sigma$ and the prediction confidence level.

\subsection{ILSVRC-2012}
\label{sec:ilsvrc}

\textbf{Comparison with previous work.}
In Table~\ref{tbl:imagenet_comparison} we compare our method with previous works.
%In Table~\ref{tbl:imagenet_comparison} we compare our method with the work of Figurnov et al. \cite{figurnov2016perforatedcnns}, Kim et al. \cite{kim2018mosaic}, Shomron and Weiser \cite{shomron2019spatial}, and Dong et al. \cite{dong2017more}.
Shomron and Weiser \cite{shomron2019spatial}, and Kim et al. \cite{kim2018mosaic} focus on predicting zero-valued activations according to nearby zero-valued activations.
Figurnov et al. \cite{figurnov2016perforatedcnns}, on the other hand, mainly use ``nearest neighbor'' to interpolate missing activation values.
Dong et al. \cite{dong2017more} present a network structure with embedded low-cost convolution layers that act as zero-valued activation masks for the original convolution masks.

Using AlexNet and VGG-16, our method shows better results across the board (besides a small difference of 0.07 with VGG-16 top5 compared to Kim et al.) --- for more MAC savings, a smaller accuracy drop is observed.
%for the same accuracy, more MAC savings are achieved, and for the same MAC savings, a smaller accuracy drop is observed.
Regrading ResNet-18, our method shows better results compared with Shomron and Weiser but falls short when compared with Dong et al.
Notice though that the method Dong et al. introduced involves an entire network training with \emph{hundreds} of epochs.
Therefore, even though there is some resemblance between our method and theirs, the results themselves are not necessarily comparable. 

All MAC saving measurements reported in Table~\ref{tbl:imagenet_comparison} are theoretical, that is, they are not actual speedups.
However, it is apparent that the \emph{potential} of \pred{} is greater than the other previous works.
We discuss related hardware implementations and related work next.

\begin{table}[t]
\caption{MAC reduction and accuracy with and without fine-tuning (ft) compared with previous work.
		 Thresholds are set by solving Equation~\ref{eq:non-linear} (see supplementary material for execution details).
		 The accuracy columns represent the decrease in accuracy.
         The MAC columns represent the MAC operations reduction.
         The minus ('-') symbol represents unpublished data.}

\centering
\begin{tabular}{l|l|cc|cc|c|cc|cc|c}
\hline
\multicolumn{1}{c|}{} & \multicolumn{6}{c|}{Related Work} & \multicolumn{5}{c}{Ours} \\
\multicolumn{1}{c|}{\multirow{-2}{*}{Net}} & \multicolumn{1}{c|}{Paper} & \,Top1\, & \,ft\, & \,Top5\, & \,ft\, & \,MAC\, & \,Top1\, & \,ft\, & \,Top5\, & \,ft\, & \,MAC\, \\ \hline
 & Figurnov et al. & - & - & 8.50 & 2.0 & 50.0\% & 4.63 & 1.97 & \cellcolor[HTML]{D4D4D4}3.04 & \cellcolor[HTML]{D4D4D4}1.17 & \cellcolor[HTML]{D4D4D4}51.1\% \\
 & Kim et al. & 0.48 & - & 0.40 & - & 28.6\% & \cellcolor[HTML]{D4D4D4}0.34 & 0.33 & \cellcolor[HTML]{D4D4D4}0.24 & 0.14 & \cellcolor[HTML]{D4D4D4}32.4\% \\
\multirow{-3}{*}{AlexNet} & Shomron et al. & 4.00 & 1.6 & 2.90 & 1.3 & 37.8\% & \cellcolor[HTML]{D4D4D4}1.28 & \cellcolor[HTML]{D4D4D4}0.78 & \cellcolor[HTML]{D4D4D4}0.84 & \cellcolor[HTML]{D4D4D4}0.42 & \cellcolor[HTML]{D4D4D4}38.0\% \\ \cline{1-7}
 & Figurnov et al. & - & - & 15.6 & 1.1 & 44.4\% & 11.02 & 1.27 & \cellcolor[HTML]{D4D4D4}7.02 & \cellcolor[HTML]{D4D4D4}0.71 & \cellcolor[HTML]{D4D4D4}44.5\% \\
 & Kim et al. & 0.68 & - & \cellcolor[HTML]{D4D4D4}0.26 & - & 25.7\% & \cellcolor[HTML]{D4D4D4}0.54 & 0.24 & 0.33 & 0.11 & \cellcolor[HTML]{D4D4D4}26.2\% \\
\multirow{-3}{*}{VGG-16} & Shomron et al. & 3.60 & 0.7 & 2.00 & 0.4 & 30.7\% & \cellcolor[HTML]{D4D4D4}1.71 & \cellcolor[HTML]{D4D4D4}0.31 & \cellcolor[HTML]{D4D4D4}0.87 & \cellcolor[HTML]{D4D4D4}0.19 & \cellcolor[HTML]{D4D4D4}31.3\% \\ \hline
 & Dong et al. * & - & \cellcolor[HTML]{D4D4D4}3.6 & - & \cellcolor[HTML]{D4D4D4}2.3 & \cellcolor[HTML]{D4D4D4}34.6\% & 12.35 & 7.22 & 8.37 & 4.66 & 34.0\% \\
\multirow{-2}{*}{ResNet-18} & Shomron et al. & 11.0 & 2.7 & 7.60 & 1.7 & 22.7\% & \cellcolor[HTML]{D4D4D4}2.96 & \cellcolor[HTML]{D4D4D4}1.86 & \cellcolor[HTML]{D4D4D4}1.70 & \cellcolor[HTML]{D4D4D4}1.13 & \cellcolor[HTML]{D4D4D4}23.8\% \\ \hline
\end{tabular}
\label{tbl:imagenet_comparison}
\end{table}

\section{Discussion and Related Work}
\label{sec:related_work}
\textbf{Hardware.} It is not trivial to attain true benefits from mainstream compute engines, such as GPUs, when dealing with compute intensive tasks, such as CNNs, and using a conditional compute paradigm of which only a portion of the MAC operations are conducted and the rest are performed or skipped according to the previously computed results.
However, the implementation of CNN \emph{accelerators}, which are capable of doing so and gain performance, has already been demonstrated.
Kim et al. \cite{kim2018mosaic} present an architecture based on SCNN \cite{parashar2017scnn} which is capable of skipping computations of entire ofm activations based on already computed zero-valued ofm activations.
Moreover, Hua et al. \cite{hua2019boosting}, Akhlaghi et al. \cite{akhlaghi2018snapea} and Asadikouhanjani et al. \cite{asadikouhanjani2020novel} propose hardware that is capable of saving a portion of the MAC operations needed per ofm activation based on the accumulated partial sum. Akhlaghi et al. also suggest a method to do so in a speculative manner.

\textbf{Speculative execution.}
GPPs make extensive use of speculative execution \cite{hennessy2011computer}.
%For example, branch predictors are used to predict whether a branch instruction will be taken (or not taken) based on previous encounters with that branch or other nearby branches, and caches are used to store values that will probably be accessed repetitively in the near future.
They leverage unique application characteristics, such as code semantics and temporal locality, to better utilize the GPP's internal structure to achieve better performance.
Researchers have also studied the speculative execution of CNNs to decrease their compute demands.
%One popular approach is via sign prediction which is made possible thanks to the ReLU activation function \cite{nair2010rectified}.
%Akhlaghi et al. \cite{akhlaghi2018snapea} predict during the convolution computation whether the convolution results will end up negative.
Song et al. \cite{song2018prediction}, Lin et al. \cite{lin2017predictivenet}, and Chang et al. \cite{chang2018reducing} predict whether an entire convolution result is negative according to a partial result yielded by the input MSB bits.
Huan et al. \cite{huan2016multiplication} avoid convolution multiplications by predicting and skipping near-zero valued data, given certain thresholds.
Chen et al. \cite{chen2018efficient} predict future weight updates during training based on Momentum \cite{qian1999momentum} parameters.
Zhu et al. \cite{zhu2018sparsenn} use a fully connected layer to predict each ofm activation sign and low-rank approximation to decrease the weight matrix.

%In conventional GPPs, control and data mispredictions may lead to a pipeline flush and a rewind of the program counter, i.e., it is common that an execution stops and starts over due to mispredictions.
%Interestingly, this is not the case with DNNs, since they are statistics-based and tolerant to errors (to some extent) \cite{li2017understanding}.
%Therefore, the impact of mispredictions on the model accuracy depends on \emph{how many} non-zero-valued activations are predicted as zero, as well as on \emph{which} values were zeroed out.

\textbf{Spatial correlation.} 
The correlation between neighboring activations in CNN feature maps is an inherent CNN characteristic that may be exploited.
The works by Shomron and Weiser \cite{shomron2019spatial}, Kim et al. \cite{kim2018mosaic}, and Figurnov et al. \cite{figurnov2016perforatedcnns} were described at Section~\ref{sec:ilsvrc}.
In addition, Kligvasser et al. \cite{kligvasser2018xunit} propose nonlinear activations with learnable spatial connection to enable the network capture more complex features, and Mahmoud et al. \cite{mahmoud2018diffy} operate on reduced precision deltas between adjacent activations rather than on true values.

\textbf{Dynamic pruning.}
As opposed to static pruning \cite{han2015deep}\cite{hu2016network}\cite{luo2017thinet}\cite{luo2017thinet}\cite{li2016pruning}, dynamic pruning is input-dependent. %, \ie, part of the computations may be skipped depending on the current inputs.
Dong et al. \cite{dong2017more} present a network structure with embedded low-cost convolution layers that act as zero-valued activation masks for the original convolution masks.
In a higher granularity, Lin et al. \cite{lin2017runtime} use reinforcement learning for channel pruning,
%Hua et al. \cite{hua2019boosting} use a network structure with embedded channel gating capability,
Gao et al. \cite{gao2018dynamic} prune channels according to a channel saliency map followed by a fully connected layer, and He et al. \cite{he2017channel} propose a two-step channel pruning algorithm with LASSO regression and fine-tuning.
All these works, except \cite{he2017channel}, involve extensive model training.

%\textbf{Network for network.} Conceptually, our work demonstrates how a NN is used for the benefit of another NN and so it may relate to neural architecture search (NAS) work, a field in which model architectures are formed automatically by another network \cite{zoph2016neural, tan2019mnasnet}.
%In addition, some works on dynamic pruning also share this concept \cite{gao2018dynamic, lin2017runtime}, as mentioned previously.

Our work is most closely related to the work of Dong et al., Shomron and Weiser, Kim et al., and Figurnov et al., with which we have also compared our results.
However, our work provides the user with a continuous range of operating points and offers a prior estimate of the model accuracy degradation and MAC savings.
Specifically, in contrast to Dong et al., our approach does not require labeled data (yet, it can be used for fine-tuning) and does not require hundreds of epochs for training.
As for Shomron and Weiser, Kim et al., and Figurnov et al., we create the binary prediction masks using a CNN-based approach.

%As opposed to Dong et al., our approach is easy to deploy, \ie, it does not require training of the entire model and it offers the user many operating points to choose from.
%In addition, our prediction mask relies on partial ofm computation rather than on the ifm.
%As for Shomron and Weiser, Kim et al., and Figurnov et al., we create binary prediction masks using a CNN-based approach and offer an estimate of the model accuracy degradation and MAC savings.

\section{Conclusions}
We propose a zero activation predictor (\pred{}) that dynamically identifies the zero-valued output feature map (ofm) activations prior to their computation, thereby saving their convolution operations.
\pred{}	exploits the spatial correlation of ofm activations inherent in convolution neural networks, meaning that according to a sparsely computed ofm, \pred{} determines whether the remaining activations are zero-valued or non-zero-valued.
\pred{} is a lightweight CNN that imposes negligible computation and parameter overheads and its deployment and training does not require labeled data or modification of the baseline model architecture and parameters.
In addition, \pred{} speculation level is tunable, allowing an efficient \textit{a priori} control of its accuracy-savings trade-off.

\subsubsection*{Acknowledgments}
We acknowledge the support of NVIDIA Corporation for its donation of a Titan V GPU used for this research.

\bibliographystyle{splncs04}
\bibliography{egbib}
\end{document}